# Addressing Uncertainty in Imbalanced Histopathology Image Classification of HER2 Breast Cancer: An interpretable Ensemble Approach with Threshold Filtered Single Instance Evaluation (SIE)

MD SAKIB HOSSAIN SHOVON[1], M. F. MRIDHA[1](Senior Member, IEEE), KHAN MD HASIB[2]( Member, IEEE), SULTAN ALFARHOOD[3], MEJDL SAFRAN[3], AND DUNREN CHE[4]
[1]Department of Computer Science, American International University-Bangladesh, Dhaka 1229, Bangladesh
[2]Department of Computer Science and Engineering, Bangladesh University of Business and Technology, Dhaka 1216, Bangladesh
[3]Department of Computer Science, College of Computer and Information Sciences, King Saud University, P.O.Box 51178, Riyadh 11543, Saudi Arabia; (e-mail: sultanf@ksu.edu.sa; mejdl@ksu.edu.sa)
[4]School of Computing, Southern Illinois University, Carbondale, IL 62901, USA;(e-mail: dche@cs.siu.edu)

Corresponding author: M. F. MRIDHA(firoz.mridha@aiub.edu), MEJDL SAFRAN(mejdl@ksu.edu.sa).

This research is funded by the Researchers Supporting Project Number (RSPD2023R1027), King Saud University, Riyadh, Saudi Arabia.

**ABSTRACT**
Breast Cancer (BC) is among women's most lethal health concerns. Early diagnosis can alleviate the mortality rate by helping patients make efficient treatment decisions. Human Epidermal Growth Factor Receptor (HER2) has become one the most lethal subtype of BC. According to the College of American Pathologists/American Society of Clinical Oncology (CAP/ASCO), the severity level of HER2 expression can be classified between 0 and 3+ range. HER2 can be detected effectively from immunohistochemical (IHC) and, hematoxylin & eosin (HE) images of different classes such as 0, 1+, 2+, and 3+. An ensemble approach integrated with threshold filtered single instance evaluation (SIE) technique has been proposed in this study to diagnose BC from the multi-categorical expression of HER2 subtypes. Initially, DenseNet201 and Xception have been ensembled into a single classifier as feature extractors with an effective combination of global average pooling, dropout layer, dense layer with a swish activation function, and l2 regularizer, batch normalization, etc. After that, extracted features has been processed through single instance evaluation (SIE) to determine different confidence levels and adjust decision boundary among the imbalanced classes. This study has been conducted on the BC immunohistochemical (BCI) dataset, which is classified by pathologists into four stages of HER2 BC. This proposed approach known as DenseNet201-Xception-SIE with a threshold value of 0.7 surpassed all other existing state-of-art models with an accuracy of 97.12%, precision of 97.15%, and recall of 97.68% on H&E data and, accuracy of 97.56%, precision of 97.57%, and recall of 98.00% on IHC data respectively, maintaining momentous improvement. Finally, Grad-CAM and Guided Grad-CAM have been employed in this study to interpret, how TL-based model works on the histopathology dataset and make decisions from the data.

## I. INTRODUCTION

**B**REAST cancer (BC) is a fatal disease in which cells grow abnormally and spread uncontrollably in different breast parts. It is the most common cancer in women, estimated with 1958310 new cases and 609820 deaths annually around the United States in 2023 [1]. Based on molecular expressions, characteristics of BC can be classified into four subtypes such as 1. Luminal A, 2. Luminal B, 3. Human

Epidermal Growth Factor Receptor (HER2), 4. Basal-like. Among these, HER2 is the most damning subtype [2]. According to a study, it is amplified and/or over-expressed in about 25% of BC cases [3].

Congenial diagnosis in the early stage can reduce mortality from BC by 40% [4]. To diagnose BC and determine its status, making precise decisions is very important by analyzing the characteristics of specific proteins such as HER2. Histopathological image from biopsy test is considered the standard gold technique for identifying BC compared to other medical imaging methods; for example, Mammography, Magnetic Resonance Imaging (MRI), Ultrasound, and Computed Tomography (CT). This histopathological image has been potential to recognize different types and stages of BC [5]. Unfortunately, this biopsy test brings a great challenge to maintain standard practice because of requiring specialized pathologists, laboratories, expensive reagents, equipment & additional time. Therefore, these limitations should be mitigated by finding a more effective way to diagnose BC.

Pathologists examine IHC-stained slides to determine HER2 status. It is used not only to define the cancer cell's receptors but also to characterize various subtypes of tumors and provides essential information about post-treatment [6]. Per the present direction by the College of American Pathologists American Society of Clinical Oncology (CAP/ASCO), define the score of HER2 between 0-3+. HER2 negative (HER2-) is classified as a score of 0 or 1+. HER2 positive (HER2+) is classified as a score of 3+, and a score of 2+ is categorized as the obscure expression [7]. Some medicines, for example, Trastuzumab, Lapatinib, etc., are available to reduce the death rate though these drugs are not only worthless but also detrimental in some cases. Hence, digital pathology techniques can be investigated by incorporating AI and Machine Learning to mitigate all these issues concerning time, cost, and accurate detection of BC.

In our research, TL based enhanced ensemble model was picked as a feature extractor on the H&E & IHC- stained data. According to a study, TL has been applied for the segmentation, detection, and classifying, of US BC images to develop a comparatively better performing model for BC research in the early stage [8]. Furthermore, a lot of recent research work has been obtained using the TL method in Mammographic Breast Mass classification [9], BC classification on Histopathological images [10], Ultrasound BC image classification [11], etc.

Accomplishing the advancement of TL in recent trends, CNN based approach has been conducted in this study. However, having no research dealing with CNN-based data imbalanced issues of the dataset in digital pathology, a new threshold-based SIE technique has been integrated with this study to interpret the dataset of BC. This paper represents an ensemble approach with threshold-filtered single instance evaluation to handle uncertainty among different imbalanced datasets. We trained our model on H&E and IHC images from the BCI dataset to predict the multistage of BC among different classes to explain confidence scores deeply, guided by ASCO & CAP. Finally, we explained our model using Grad-CAM and Guided Grad-CAM.

The following are the principal contributions of this study:

- Initially, several state-of-the-art models have been experimented. Among these, DenseNet201 & Xception performed best. Hence, An enhanced ensemble approach has been introduced based on DenseNet201 and Xception as a feature extractor with a combination of global average pooling, dropout mechanism, dense layer with swish activation function and l2 regularizer, and batch normalization layer.
- Extracted information has been passed through single instance evaluation (SIE) by setting up different threshold values. By incorporating this technique model interprets its performances in different confidence scores. The model excludes extremely low confidence scores and makes predictions based on existing data. It helps to reduce noise and address challenges associated with imbalanced classes.
- Based on filtered data model calculates accuracy, weighted average precision, and macro average recall which also contributes to boosting classification performances. The model output obtained the best performances with a threshold value of 0.7 and surpassed all other existing works maintaining momentous improvement.
- Finally, Grad-CAM and Guided Grad-CAM have been utilized for different classes of BC to ensure how our model works for this particular complex dataset. Guided Grad-CAM outperformed Grad-CAM slightly by obtaining more intuitive visualization.

The following section is organized as follows. In Section (2) related work has been described. In section (3), materials and methods have been discussed. Section (4) describes the result analysis. Section (5) illuminates challenges and future scopes. Finally, section (6) concludes the conclusion of the research.

## II. RELATED WORK

In the context of digital pathology, immense interest is being grown using deep learning (DL) techniques to support the decision-making of pathologists dealing with cancer-related topics [12]. Various imaging methods are being used to identify BC, such as Mammograms (MGs) [13], Histopathology (HP) [14], Magnetic Resonance Imaging (MRI) [15], Computed Tomography (CT) [16], Ultrasounds (US) [17], etc. As the histopathological images from the biopsy test are defined as the gold standard procedures to diagnose BC compared to other techniques; in this study, Immunohistopathology (IHC) and hematoxylin and eosin (H&E) stained images have been utilized to detect multi-stages of BC from HER2 expression.

There has been significant research to diagnose BC on histopathology images. DL has been applied heavily to analyze tumor detection, grading, subtypes, predictions, etc [18]. Lin et al. [19] acquired 77.50% accuracy using single-layer CNN. Darken et al. [20] proposed a better LeNet-5 (RMSprop) method and achieved 82.58% accuracy. Togacar et al. [21] gained a better result of 83.19% accuracy using CNN with the Taguchi method. After that, Cheng demonstrated a more robust model, a Uniform Experimental Design (UED), and optimized the CNN parameters of histopathological image classification of BC that surpassed all previous results with 84.41% accuracy. Later on, many studies were conducted for various tasks like detection, classification, and multi-classification of histopathology images.

Man et al. [22] proposed deep learning (DL) model with handcrafted features for mitosis detection using BreaKHis dataset and acquired promising results of 92% Precision, 88% Recall, and 90% F-Score. Another study by Perumal et al. [23] was conducted for the multi-classification task using Hybrid CNN+ Deep RNN on BACH(ICIAR 2018) dataset and obtained satisfying accuracy of 91.3%. In addition, different techniques have been demonstrated for BC classification. Rulaningtyas R et al. [24] achieved 88% accuracy applying GLCM and Neural Network method. Spanhol et al. [25] obtained 81-85% accuracy using the CaffeNet model on the BreaKHis dataset for eight subtypes. Nahid et al. [26] proposed different machine learning techniques and DL methods (K-Means and Mean-Shift clustering as a feature extractor, CNN, and LSTM structure on the same dataset and achieved a satisfactory result of 91% accuracy.

There hasn't been significant research on the BCI dataset. Mridha et al. [27] initially conducted an experiment on the BCI dataset. They proposed TL based model called convoHER2 in their experiment. convoHER2 achieved 85.1% and 87.79% training accuracy for the H&E and IHC data respectively. Though other validation data or important matrix hasn't been shown properly. Later on, Shovon et al. [14] experimented more improved approach to the H&E image data from the BCI dataset. HE-HER2Net was proposed based on an enhanced Xception model. HE-HER2Net acquired almost an accuracy of 0.87, a precision of 0.88, and a recall of 0.86. Finally, Wang et al. [28] proposed HAHNet which combines multi-scale features with attention mechanisms and obtained better results of an accuracy of 93.65%, precision of 93.67% and recall of 92.46%, etc on the same H&E data from BCI dataset.

Every research work has been conducted to accomplish challenges such as maintaining magnification factor dependency, grade classification, class classification, accuracy, etc. Hence, more research is needed to mitigate more problems while maintaining less computational complexity and good performances in histopathology research. It is also notable to state that, no other research didn't demonstrate about data uncertainty problem. It is important to address different challenges among the imbalanced class and to identify significant confidence scores of different classes in classification problems. In addition, per our knowledge, there has not been any research to perform multi-stage classification of HER2 expression of four categories from IHC/HE data. In addition, there is no publicly available dataset without BCI, which contains four types of HER2 expression to diagnose BC. Hence, it motivates us to conduct a multi-stage classification of BC on the BCI dataset.

### III. MATERIALS AND METHODS
#### A. DATASET & PREPROCESSING

To conduct research, finding a good dataset is mandatory. Some publicly available datasets contain different classes, such as BreaKHis, BCC, BACH, TUPAC16, IDC, and BCI. Unfortunately, no other dataset without BCI is publicly available, as per our best knowledge containing four stages of HER2 BC. In our study, the BCI dataset [29] has been taken to conduct a multistage classification problem on H&E and IHC data. BCI contains 4870 registered image pairs with a resolution of 1024*1024 containing the equal number of H&E and IHC images of four distinct stages (HER2 0, HER2 1+, HER2 2+, HER2 3+). In this study, 3896 images of both H&E and IHC data from the BCI dataset have been taken for training data, and 977 images for the validation data.

For the data preprocessing step, several effective operations have been implied to reduce the overfitting problem, accelerate training time, improving the performance of the deep learning model. The primary dataset consists of RGB coefficient ranges between 0-255. However, It isn't easy to process for the model having such higher values. Eventually, rescale operation was applied by scaling with a 1/.255 factor to target the data values between 0 and 1. In addition, data has been resized to 299*299 pixels to train effectively and swiftly because the primary data was a much higher resolution of 1024*1024. In an experiment led by Sergio et al. [30], bigger image resolution always demands neural net four times the input pixels, which conspicuously requires more training time. Besides that, decent data augmentation has been implied in this research to improve the model performance. In this study, several approaches have been taken to investigate how data augmentation technique affects effectively. Heavy augmentation was unsuitable for this research as the microscopic dataset was very challenging to understand visually. Moderate augmentation process applying width and height shift within a small range of 0.1 and flipping vertically improved performance very precisely.

#### B. PROPOSED APPROACH
1) Enhanced Ensemble Model as Feature Extractor

An enhanced ensemble has been created as a feature extractor based on DenseNet201 [31] and Xception [32]. Several modifications have been included to both of the original DenseNet201 & Xception models by integrating global average pooling (GAP), one dropout function, four dense layers with swish activation function and l2 regularizers, and, two batch normalization (BN) layers. The final classification dense layer has been replaced with four neurons for the four-class multi-stage classification problem. Shovon et al. [33] conducted an experiment in which they showed momentous improvements using these optimization strategies handling underfitting or overfitting issues in their proposed ensemble network. The proposed diagram of this study is illustrated in FIGURE 1.

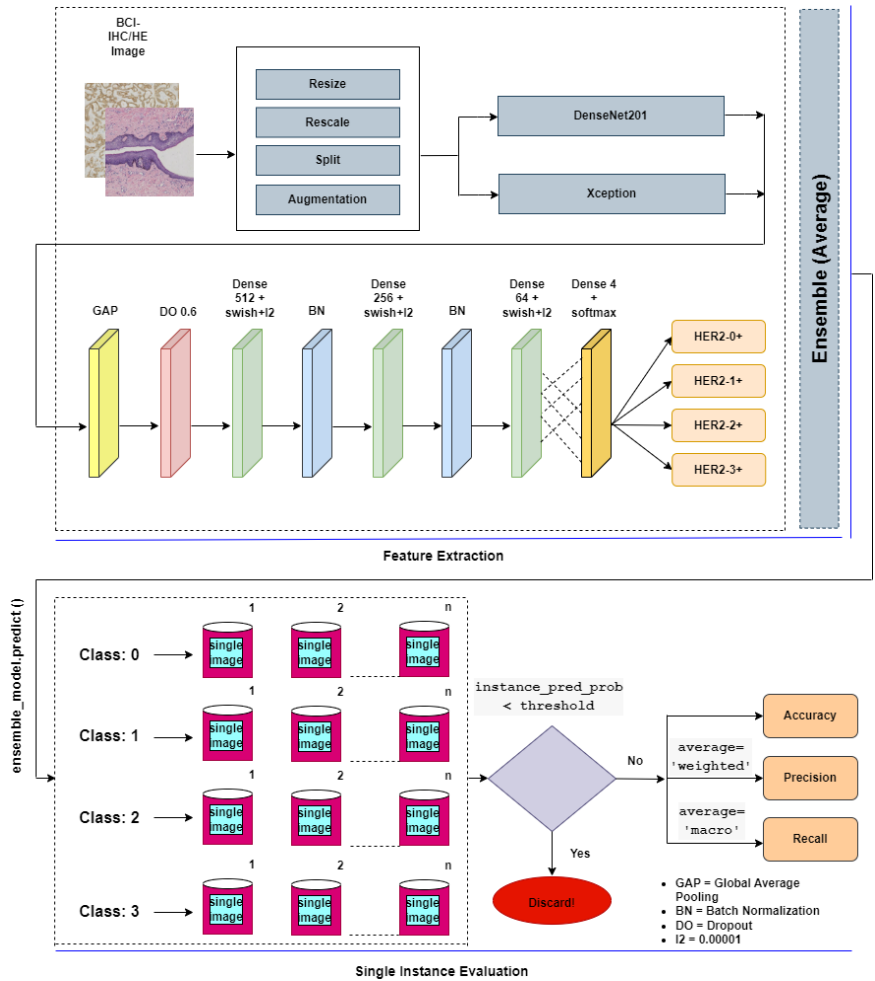

**FIGURE 1:** This figure illustrates the proposed approach of this study.

At first, the data preprocessing step was accomplished as described in the dataset & preprocessing section. Next, both Dense201 and Xception models have been modified by introducing several techniques to reduce the overfitting problem, extract more spatial information, speed up the training time, and stabilize the model more effectively. GAP has been added to the last dense layer of each of the models. GAP reduces spatial dimensions by reducing the model's parameters where a tensor of dimension h x w x d is reduced to dimension 1 x 1 x d by obtaining the average value across the h x w channel. GAP helps the model reduce overfitting issues and speeds up the training time. FIGURE 2 indicates the workflow of GAP.

The Dropout (DO) [34] regularization technique has been applied to our network to reduce the overfitting problem. DO randomly drops neurons in training time and speeds up the training time. As a result, it makes the model lite during training. Recently, it has been a more effective regularization

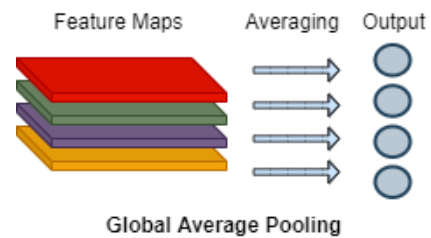

**FIGURE 2:** The workflow of Global Average Pooling

technique than others. FIGURE 3 illustrates how DO works in the neural network.

A total of three dense layers have been added in which two of which were used before each BN layer and the last one was added before the classifier layer. According to a study [35], it has been experimented, by adding more dense layers improves classification performance.

To make our more stable and faster, BN [36] has been included after the first two added dense layers. BN normalizes activation vectors from the hidden layers. Initially, it calculates mean $\mu$ and the variance $\sigma^2$ across the batch as described as follows.

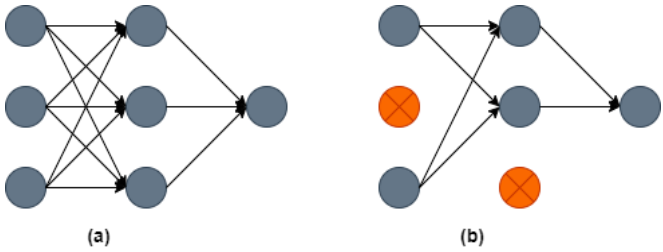

**FIGURE 3:** (a) Before applying dropout (b) After applying dropout

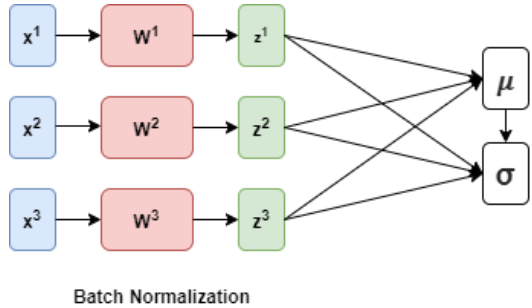

**FIGURE 4:** This figure illustrates how Batch Normalization works.

$$\mu = \frac{1}{n}\sum_i Z^{(i)} \quad (1)$$

$$\sigma^2 = \frac{1}{n}\sum_i (Z^{(i)} - \mu)^2 \quad (2)$$

After this, it normalizes the activation vector $Z^{(i)}$ as given below, where $\epsilon$ is a constant here.

$$z^{(i)}_{norm} = \frac{z_i - \mu}{\sqrt{\sigma^2 - \epsilon}} \quad (3)$$

Finally, It calculates the output $\breve{Z}$ where, $\gamma$ and $\beta$ are two trainable parameters. Here is the output formula.

$$\breve{Z} = \gamma - Z^{(i)}_{norm} + \beta \quad (4)$$

The diagram of BN is illustrated in FIGURE 4.

In the added dense layer, the Swish activation function has been mobilized with the dense layer instead of the relu function. This smooth, non-monotonic function outperforms the relu function in complex classification problems [37]. In this study, the relu and Swish functions have been experimented with, where Swish outperformed the relu function. The formula for Swish activation functions is given below.

$$Swish(x) = x\left(1 + e^{-(\beta x)}\right)^{-1} \quad (5)$$

Where $\beta$ refers to beta scalar input.

Dealing with the multiclass classification problem of BC, a dense classifier with four neurons has been applied to the end of both models with the softmax function. The mathematical intuition of softmax is described as follows.

$$\text{softmax}(z_i) = \frac{e^{z_i}}{\sum_{j=1}^{K} e^{z_j}} \quad (6)$$

Here,
$z_i$ = values of input elements,
$e^{z_i}$ = exponential function applied to $z_i$,
$\sum_{j=1}^{K} e^{z_j}$ = normalization term that indicates all the output values will sum to 1, and
K = the number of classes that belongs to the multi-class classifier.

After integrating several optimizations to both of the models, an average-based ensemble model has been created by combining them. This ensemble model generally performs better than using a single model, improves generalization, and boosts classification performances. After that, we utilized threshold-based single instance evaluation to calculate accuracy, weighted precision, and, macro recall.

2) Threshold-based Single Instance Evaluation

To understand thoroughly the problem domain & the significance of different confidence levels, it is necessary to initiate threshold-based single instance evaluation (SIE). It is applicable when the classes of the dataset are imbalanced. Sometimes, due to the uncertainty of any class in the dataset, the model can't perform classification performances properly. Eventually, it's very hard to get the proper output of the model dealing with the imbalanced dataset. In this scenario, it is important to reduce noise, evaluate the classification properly and get intuitive confidence scores for each image at different levels. In this study, after the feature extraction process through an ensemble approach, we incorporated SIE.

First, each of the images of the validation data has been predicted through the ensemble_model.predict() method of the enhanced ensemble. Predict method makes predictions of every single instance of the validation data. Next, several threshold values have been experimented with to get the confidence scores of every image. If the confidence values qualify the threshold criteria, it goes for further evaluation, otherwise, these values are excluded from evaluation. All excluded instances with extremely low-threshold values are labeled as a special label called '-1'. This is how our proposed approach reduces noises from the dataset, make adequate predictions for all the classes, and interprets the confidence scores at a different level of confidence scores. The algorithmic explanations of this approach are briefly discussed below:

1) Initialize a list of threshold values.
2) For each threshold value:
    a) Initialize empty lists for storing predictions and true labels.
    b) Iterate over the test data generator.
    c) For each batch of images and labels:

  i) Make predictions on the batch using an ensemble model.
  ii) Apply the threshold to the prediction probabilities.
  iii) Extend the prediction and true label lists with the current batch's values.
 d) Convert the prediction and true label lists to numpy arrays.
 e) Remove instances with a special label (-1) from evaluation.
 f) Calculate accuracy, precision, and recall metrics for valid predictions.
 g) Print the threshold value, accuracy, precision, and recall.
3) Repeat for each threshold value.

## C. PERFORMANCE EVALUATION

To evaluate model performance for the multi-stage classification problem; Accuracy (AC), Weighted Precision (PR), and Macro Recall (RE) have been obtained for all existing models. In our proposed approach, to boost performance metrics for the imbalanced dataset, weighted average precision has been calculated instead of simple precision, and, macro average recall has been obtained instead of simple recall. In addition, a Confusion Matrix (CM) has been visualized to demonstrate ground truth prediction vs. the predicted label prediction, where each row and column indicates predicted class and actual class instances, respectively. There are some parameters, such as True Positive (TP), False Positive (FP), True Negative (TN), and False Negative (FN), that are profoundly connected with CM. The algorithmic intuition of accuracy, weighted precision, and macro recall are described below.

**Accuracy (AC):** Accuracy is a metric that assesses the overall correctness of a model's predictions. It is determined by dividing the number of instances correctly predicted by the model by the total number of instances.

1) Initialize variables for the total correct predictions (TP), total true negatives (TN), total false positives (FP), total false negatives (FN), and the total number of instances (N).
2) For each instance:
   a) Determine the true positive (TP), true negative (TN), false positive (FP), and false negative (FN) based on the predicted label and the true label.
   b) Increment the corresponding counters accordingly.
   c) Increment the N counter.
3) Calculate the simple accuracy:
4) Divide the total number of correct predictions (TP + TN) by the total number of instances (N).
5) Multiply the result by 100 to obtain the accuracy percentage.
6) This result represents simple accuracy.

**Weighted Precision (PR):** It is a modified version of precision that considers the unequal distribution of classes in a dataset. It computes the average precision for each class, giving more weight to classes with a larger number of instances.

1) Initialize variables for the total weighted precision numerator and total weighted precision denominator.
2) For each class:
   a) Compute the precision (TP / (TP + FP)), where TP represents the number of true positives and FP represents the number of false positives for the class.
   b) Determine the weight for the class based on its number of instances.
   c) Multiply the precision by the weight and add the result to the total weighted
   d) Add the weight to the total weighted precision denominator.
3) Calculate the weighted average precision:
   a) Divide the total weighted precision numerator by the total weighted precision denominator.
   b) This result represents the weighted average precision.

**Macro Recall(RE):** It calculates the average recall across all classes without considering class imbalances. It treats each class equally and computes the recall for each class individually, then takes the average

1) For each class:
   a) Compute the recall (TP / (TP + FN)), where TP represents the number of true positives and FN represents the number of false negatives for the class.

b) Store the recall value for the class.

2) Calculate the average recall:

  a) Sum up the recall values obtained for all classes.

  b) Divide the sum by the total number of classes.

  c) This result represents the macro average recall.

When, the dataset has imbalanced class distributions, where some classes have significantly more instances than others, the weighted average method can be useful. It accounts for the class imbalance by assigning higher weights to the precisions of larger classes. This can provide a more representative measure of overall performance. On the other side, weighted average precision has been utilized to perform the balanced evaluation. Where, simple recall considers all instances equally and provides an overall measure of recall, while macro average recall treats each class equally and provides a balanced measure of recall across all classes.

### D. VISUAL EXPLANATIONS WITH GRAD-CAM & GUIDED GRAD-CAM

Building a reliable model for the users while developing a CNN-based model is essential. For the classification problem of HER2 breast cancer (BC), the Grad-CAM [38] method has been accomplished to explain our proposed model to make it more transparent. Users need to know how this model is being predicted. To clarify the model's robustness, Grad-CAM demonstrates a visual representation of the model by generating a heat map. Grad-CAM flows gradient indications to the last convolution layer of the model. However, this method is flexible enough as it can also be applied to other convolution layers. Next, Global Average Pooling is imputed to that gradient attaining neuron weights. Finally, a linear combination of the feature maps is computed, followed by a relu activation function. Thus, Grad-CAM generates a heat map of the model. In this study, Grad-CAM has been employed in the last convolution layers of the model. The formula of the Grad-CAM method is described as follows.

$$\alpha_k^c = \overbrace{\frac{1}{Z} \sum_i \sum_j}^{\text{global average pooling}} \underbrace{\frac{\partial y^c}{\partial A_{ij}^k}}_{\text{gradient via backprop}} \quad (7)$$

$$L_{\text{Grad-CAM}}^c = \text{ReLU} \underbrace{\sum_k \alpha_k^c A^k}_{\text{linear combination}} \quad (8)$$

Here,

 $y^c$ = the gradient of the score of class c,
 $A^k$ = feature map activations, and
 $\alpha_k^c$ = the weights of the neuron.

$Z$ = Number of pixels in the feature map.

In addition, Guided Grad-CAM has been experimented also to visualize the model that combines GradCAM, and GBP (Guided Backpropagation). Grad CAM is class-discriminative and locates significant image sections, and Guided Backpropagation visualizes gradients concerning the image while backpropagating through ReLU layers and highlights important pixels in the image by setting negative gradients to zero. High-resolution and class discrimination both are visualized by Guided Grad CAM. Guided GradCAM's superiority over previous techniques lies in its capacity to more thoroughly analyze and justify classification errors. With the benefit of Guided Grad CAM visualization, it is possible to clearly demonstrate what the CNN model considers while generating a prediction. This promotes transparency in deep learning models. FIGURE 5 depicts the XAI output.

### IV. RESULT ANALYSIS

To interpret the model in different confidence score levels of the imbalanced data, at first, a robust feature extractor is needed to accomplish the task efficiently. Therefore, to attain robust performance of multiclass classification problem for the IHC data, initially; all TL-based state-of-art methods such as ResNet152V2, EfficientNetB7, InceptionResNetB7, InceptionV3, ResNet101, VGG16, VGG19, Xception, and DenseNet201have been trained. As this histopathological dataset was too complex and noisy, none of the base models obtained adequate performance. ResNet152 obtained the worst performance of 67.45 % (AC), 69.01 % (PR), and 65.20% (RE) respectively while DenseNet201 derived the best performance of 81.99 % (AC), 82.69 % (PR), and 81.17 % (RE) respectively. Xception also performed closely after DenseNet201. Generally, the ensemble model performs better than an individual model. By experimenting with several ensemble combinations, DenseNet201-Xception-SIE obtained the best results compared to all other models. Eventually, we advanced to employ threshold-based single instance evaluation (SIE) with these ensemble models such as InceptionResNetV2-VGG19-SIE and DenseNet201-Xception-SIE. We experimented with different threshold values ranging from 0.3-0.7 for these models. Both of the models achieved the best performances for the threshold value of 0.7. Among these, DenseNet201-Xception-SIE (IHC) derived better results of 97.56 % (AC), 97.57 % (PR), and 98.00% (RE) respectively for the IHC data. We employed the same approach to H&E data to ensure the stability of this SIE approach. Here, DenseNet201-Xception-SIE (HE) also performed the best results compared to all other states of art models known as convoHER2, HE-HER2-Net, and HAHNet including other ensemble models which were experimented with in this study. DenseNet20-Xception-SIE (HE) surpassed the previous best model called HAHNet with an accuracy of 97.12 %, a precision of 97.15 %, and a recall of 97.68 %. As this study mainly aims to interpret the model in different confidence levels for the imbalanced dataset, there is something interesting pattern of the derived results for both datasets. If we analyze all threshold-filtered results for both of the datasets, we can observe that model acquired better results for the gradual increase of threshold values from 0.3 to 0.7. Models acquired an average of 85.00+ scores of AC, PR, and RE for both of the datasets when the threshold value is 0.3 and acquired an average of 95.00+ scores when the threshold value is 0.7. That means, most of the instances of both datasets tend to predict high confidence scores of more than 0.7 whereas few of the instances ranging between 0.3 to 0.7 have medium confidence scores. As is being seen on an average of 15.00 % + improvements when the threshold value is set to 0.3 to 0.7. TABLE 1 and TABLE 2 describes the comparative interpretation of different threshold-based approach for both datasets.

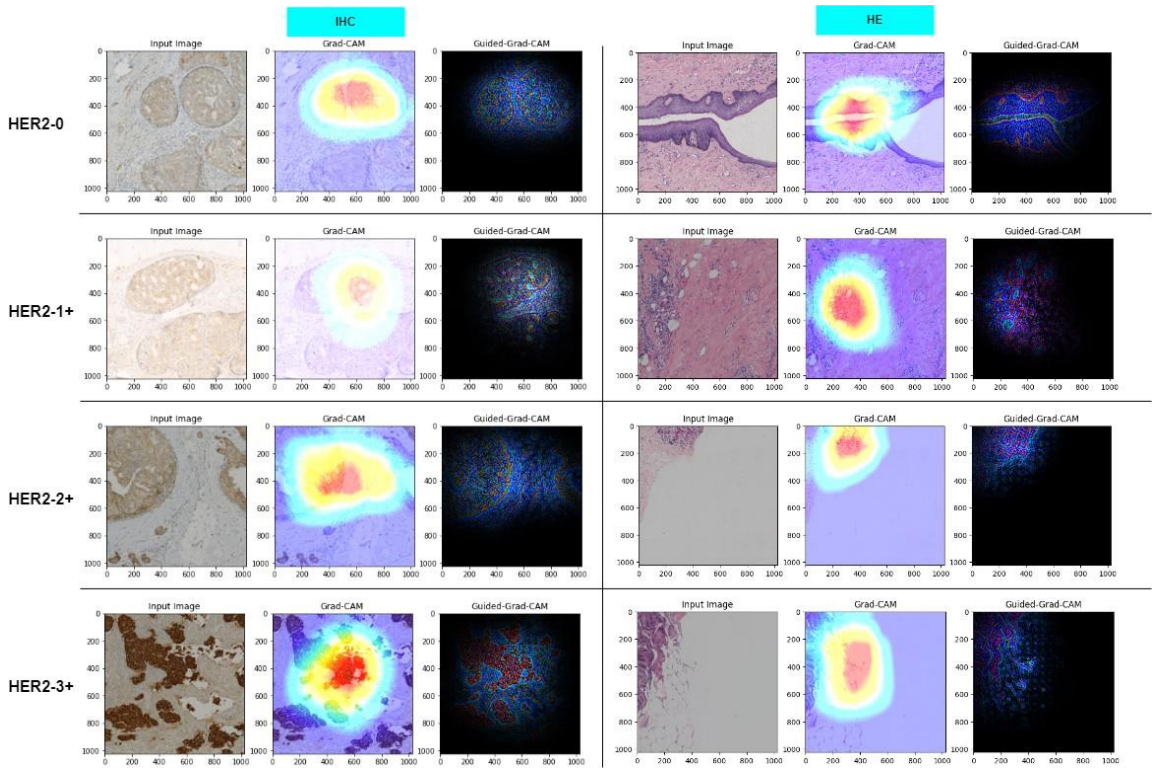

**FIGURE 5:** This figure illuminates Xai output of H&E and IHC datasets.

Handling underfitting or overfitting issues and learning efficiently, we introduced several optimization strategies such as data augmentation, global average pooling, dropout function, dense layers with swish activation function and l2 regularizer, batch normalization, etc. In addition, it is needed to incorporate a straightforward technique to handle these problems. The early stopping method was integrated into this study to handle overfitting issues. When validation loss increases consecutively for three epochs, the early stopping method automatically stops the training process. As per illustrated training and validation loss graph of H&E and IHC, we see there is not much difference between the training and validation graph that indicates the model performed well on this dataset. In addition, the validation loss graph didn't increase because of the early stop method, so overfitting never occurred in this study. FIGURE 6 indicates the training and validation graph of the H&E and IHC datasets.

We plotted a confusion matrix (CM) to elaborate true and actual level predictions among different classes. The row and the column of CM define the predicted label vs. the actual label. The figure describes the CM of H&E and IHC respectively from left to right. The diagonal deep-colored line indicates the correct predictions of our approach. On the left side of the graph, we see, only 21 predictions were wrongly classified in H&E data, and on the right side, only 20 predictions were incorrectly classified in IHC data of the validation dataset. Though these misclassified data met the threshold conditions but failed to classify correctly. This minimum number of misclassified data shows the effectiveness of this threshold-based interpretation. FIGURE 7 elaborates Confusion Matrix.

In this research, we accomplished heatmap and line plot visualization to describe model performances precisely for each evaluation metric for the different threshold values. The figure illustrates heatmap and line plot visualization respectively. The yellow-colored column at the right side of each heatmap indicates the best values obtained from the model. We see, for both datasets our proposed approach called, DenseNet201-Xception-SIE (HE) and, DenseNet201-Xception-SIE (IHC) obtained the best output values for the threshold 0.7. In addition, a line plot has been introduced to elaborate more on this study. Similarly, in the line plot, we observe the highest upward points were indicated when the threshold value was 0.7. In conclusion, according to the observation from both plots, we can say the SIE approach performed gradually better with the increasing threshold values for both of the datasets. The visualization of the heatmap and line plot are presented in FIGURE 8 and FIGURE 9 respectively.

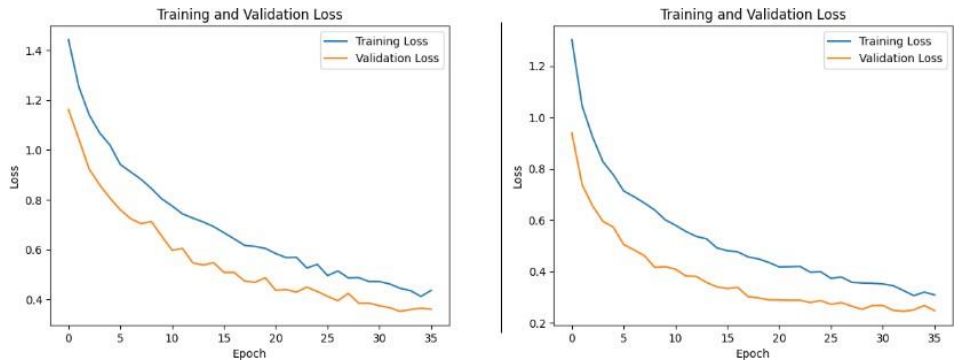

**FIGURE 6:** This figure represents the training and validation Loss graph of the H&E and IHC dataset respectively (left to right).

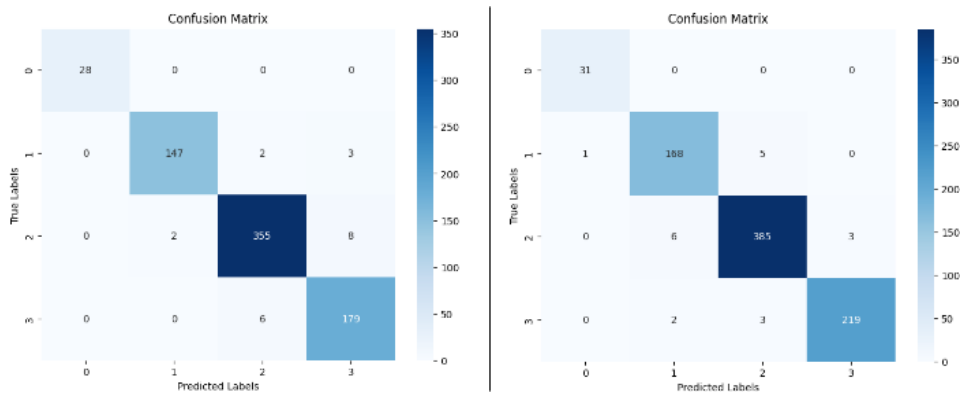

**FIGURE 7:** Explanations of the Confusion Matrix of H&E and IHC dataset (left to right).

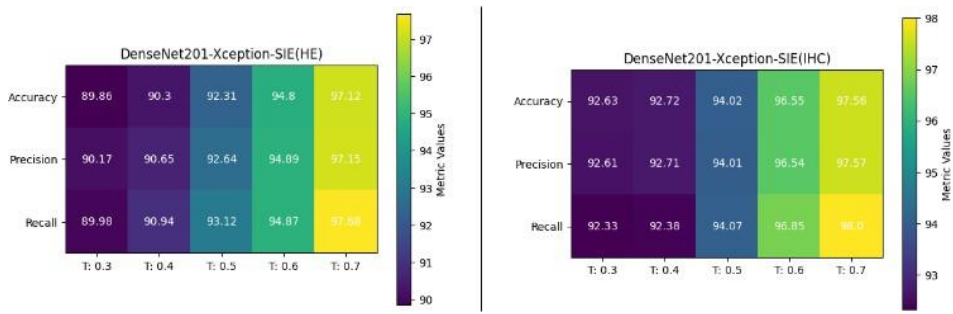

**FIGURE 8:** The given figure illustrates the Heat Map of the H&E and IHC datasets respectively.

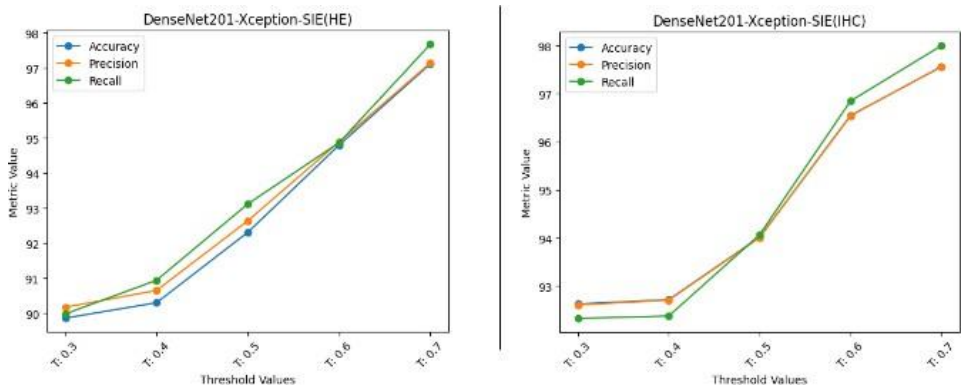

**FIGURE 9:** The illustrated figure represents Line Plots of H&E and IHC datasets respectively.

TABLE 1: This table displays the experimental results of several CNN models for the IHC data.

| Model | Accuracy (%) | Precision (%) | Recall (%) |
|---|---|---|---|
| convoHER2 (Mridha et al. [27]) | 87.79 | | |
| ResNet152V2 | 67.45 | 69.01 | 65.20 |
| EfficientNetB7 | 71.34 | 72.78 | 67.86 |
| InceptionResNetV2 | 71.75 | 75.14 | 68.37 |
| InceptionV3 | 73.39 | 75.00 | 71.55 |
| ResNet101 | 74.82 | 75.97 | 73.80 |
| VGG16 | 76.36 | 78.91 | 74.31 |
| VGG19 | 78.20 | 80.43 | 76.14 |
| Xception | 80.76 | 82.20 | 79.43 |
| DenseNet201 | 81.99 | 82.69 | 81.17 |
| InceptionResNetV2-VGG19-SIE(IHC) [Threshold: 0.3] | 85.97 | 87.34 | 87.23 |
| InceptionResNetV2-VGG19-SIE(IHC) [Threshold: 0.4] | 86.86 | 88.12 | 88.08 |
| InceptionResNetV2-VGG19-SIE(IHC) [Threshold: 0.5] | 89.29 | 90.13 | 90.53 |
| InceptionResNetV2-VGG19-SIE(IHC) [Threshold: 0.6] | 96.30 | 96.24 | 71.49 |
| InceptionResNetV2-VGG19-SIE(IHC) [Threshold: 0.7] | 96.53 | 94.18 | 49.34 |
| DenseNet201-Xception-SIE(IHC) [Threshold: 0.3] | 92.63 | 92.61 | 92.33 |
| DenseNet201-Xception-SIE(IHC) [Threshold: 0.4] | 92.72 | 92.71 | 92.38 |
| DenseNet201-Xception-SIE(IHC) [Threshold: 0.5] | 94.02 | 94.01 | 94.07 |
| DenseNet201-Xception-SIE(IHC) [Threshold: 0.6] | 96.55 | 96.54 | 96.85 |
| **DenseNet201-Xception-SIE(IHC) [Threshold: 0.7]** | **97.56** | **97.57** | **98.00** |

Finally, after attaining the best result with, DenseNet201- Xception-SIE, the Grad-CAM, and Guided Grad-CAM methods have been used based on Xception to interpret the model in the last convolution layer of block14_sepconv2_act. For the imbalanced dataset, it is essential to know how the model learns and makes predictions from the data. One image from each class of the H&E and IHC datasets has been taken to experiment with these methods to leverage the model learned in the last convolution layer. As per the author's suggestion, Grad-CAM generates the most precise heat map in the last convolution layer as the CNN model obtains more spatial information. Eventually, the visualization of the last convolution layer (block14_sepconv2_act) illustrated the precise outputs from these images. The bright area of the image indicates that our model predicted breast cancer, particularly based on that area. As the components of IHC images are microscopic, Grad-CAM generalizes all the potential areas aggregately by producing a heat map. Many areas of the images have been highlighted. This shows how the model focused on certain image regions that were emphasized while classifying numerous input images. On the other side, Guided Grad-CAM worked more precisely in terms of plotting contours and edges more sharply. If it is compared to the outputs of Grad-CAM and Guided Grad-CAM; almost all of the classes Guided Grad-CAM focused more precisely rather than plotting a general circle, on the main region of the IHC images. From the graphic illustrations of Grad-CAM and Guided Grad-CAM analysis, we can observe that Grad-CAM detects the primary areas altogether, but Guided Grad-CAM ascertains regions more precisely.

TABLE 2: This table displays the experimental results of several CNN models for the H&E data.

| Model | Accuracy (%) | Precision (%) | Recall (%) |
|---|---|---|---|
| convoHER2 (Mridha et al. [27]) | 85.10 | | |
| HE-HER2Net ( [14] | 87.00 | 88.00 | 86.00 |
| HAHNet (Wang et al. [28]) | 93.65 | 93.67 | 92.46 |
| InceptionResNetV2-VGG19-SIE(HE) [Threshold: 0.3] | 78.49 | 79.37 | 79.07 |
| InceptionResNetV2-VGG19-SIE(HE) [Threshold: 0.4] | 80.51 | 81.11 | 81.66 |
| InceptionResNetV2-VGG19-SIE(HE) [Threshold: 0.5] | 84.27 | 84.68 | 85.15 |
| InceptionResNetV2-VGG19-SIE(HE) [Threshold: 0.6] | 89.40 | 89.01 | 62.87 |
| InceptionResNetV2-VGG19-SIE(HE) [Threshold: 0.7] | 95.78 | 96.46 | 95.38 |
| DenseNet201-Xception-SIE(HE) [Threshold: 0.3] | 89.86 | 90.17 | 89.98 |
| DenseNet201-Xception-SIE(HE) [Threshold: 0.4] | 90.30 | 90.65 | 90.94 |
| DenseNet201-Xception-SIE(HE) [Threshold: 0.5] | 92.31 | 92.64 | 93.12 |
| DenseNet201-Xception-SIE(HE) [Threshold: 0.6] | 94.80 | 94.89 | 94.87 |
| **DenseNet201-Xception-SIE(HE) [Threshold: 0.7]** | **97.12** | **97.15** | **97.68** |

Moreover, It is important to maintain the ethical and regulatory aspects of implementing AI for medical diagnosis, especially in BC. This is because AI is a powerful technology that can have a significant impact on people's lives, and it is important to ensure that it is used in a responsible and ethical way. One important issue is the biases of the data. As AI algorithms are trained on data, it is important to ensure that, this data is representative of the population that the AI system will be used to diagnose. If the data is biased, then the AI system will also be biased, and this could lead to inaccurate diagnoses. In this study, benchmark datasets were experimented with which datasets were labeled by expert pathologists to ensure the quality of the data. Another important concern is model transparency. It is important to be transparent about how AI systems work and how they make decisions. This is important for patients, who need to be able to understand how the AI system is being used to diagnose them, and it is also important for pathologists, who need to be able to trust the AI system's recommendations. To address this concern in this research, the xai interpretation has been utilized. To sum up, our research addresses all ethical concerns to provide an AI solution for the BC.

## V. CHALLENGES AND FUTURE SCOPES

There were some challenging issues to accomplish this study. As this dataset is noisy, it's very hard to differentiate the instances among different classes visually. That's why almost every tl-based existing model couldn't perform well on this dataset. To initialize threshold-based single instance evaluation, feature extraction is an unavoidable step. We employed an enhanced ensemble method for that task. However, an ensemble method is more complex than a single tl-based model, therefore, it requires more complex parameters and takes more training time. In addition, as per the training and validation loss graph, our proposed model is slightly underfitted though it has no great impact on this study. There is another limitation that belongs to our study and that is information loss during the evaluation process. Because extremely low confidence predicted instances have been excluded from the evaluation this occurs some information loss. In addition, it is not standard practice to compare model performances to general tl-based output. That's why, we incorporated a threshold-based approach and evaluated based on the threshold filtered data.

However, there are some scopes to work on ex: more optimization techniques can be integrated to handle overfitting issues more precisely. More threshold values can be explored to get more promising results. In addition, an attention mechanism can be introduced to elaborate more intuitive visual output through an explainable method. Grad-CAM++ and Score-CAM can be incorporated as these explainable methods are being used vastly. These methods are showing promising results in terms of complex data.

## VI. CONCLUSION

HER2 subtypes have been the most venturesome health issues in breast cancer (BC) disease. Prior diagnosis can help pathologists reduce death rates by taking appropriate patient care. There are being a lot of research to diagnose breast cancer though there is hardly any experiment to handle internal uncertainty issues of the dataset. Histopathology images of different stage of BC are very hard to classify as these data looks very similar to differentiate. Hence, our study aims to solve data imbalanced issues, reduce noise from the data, and perform threshold-based confidence level interpretation. Initially, an enhanced ensemble model based on DenseNet201 and Xception has been introduced with several optimization techniques. This enhanced ensemble robustly extracts features from images compared to all other existing models. Next, threshold-based single instance evaluation (SIE) has been incorporated to treat every image individually and make in-depth interpretations. Our threshold-based approach with a value of 0.7 surpassed all other state-of-art model performances significantly. Moreover, it interprets confidence scores of different levels of the imbalanced dataset. Our approach also incorporates effective techniques by removing extremely low-confidence predictions from evaluation. Finally, accuracy, weighted average precision, and macro average recall have been obtained to boost classification performances during evaluation. In addition, Grad-CAM and Guided Grad-CAM precisely explain images of distinct classes for both datasets by generating a heatmap that helps to understand model performances for the imbalanced dataset. Overall, this research introduces diverse research directions that can be directed to histopathological BC research in digital pathology so that pathologists and researchers can efficiently make decisions, learn about data, and diagnoses efficiently.

## ACKNOWLEDGEMENT

The authors extend their appreciation to King Saud University for funding this research through Researchers Supporting Project Number (RSPD2023R1027), King Saud University, Riyadh, Saudi Arabia.

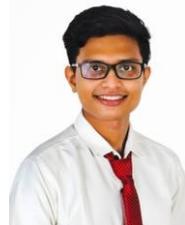

**MD SAKIB HOSSAIN SHOVON** received a B.Sc. degree in computer science and engineering (CSE) with a major in information systems from American International University-Bangladesh (AIUB) in 2023. He was a Research Assistant (RA) at AIUB (May 2022 – August 2022). He works at Advanced Machine Intelligence Research Lab (AMIRL) as a Research Co-ordinator & Lead Research Assistant (July 2022 – Continuing). In addition, he is a trainee Machine Learning (ML) Engineer under the Government Edge Project associated with HeadBlocks (28th May 2023 – ongoing). He has published many articles in prestigious journals. His research interests include Classical Machine Learning (CML), Quantum Machine Learning (QML), Federated Machine Learning (FML), Reinforcement Learning (RL), Computer Vision (CV), Natural Language Processing (NLP), Recurrent Neural Network (RNN) and Explainable AI (XAI).

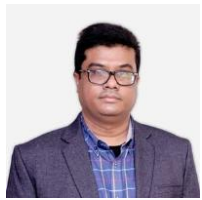

**DR. M. F. MRIDHA** (Senior Member, IEEE) received the Ph.D. degree in AI/ML from Jahangirnagar University, in 2017. He is currently working as an Associate Professor with the Department of Computer Science, American International University-Bangladesh (AIUB). Before that, he worked as an Associate Professor and the Chairperson at the Department of CSE, Bangladesh University of Business and Technology. He also worked as the CSE Department Faculty Member at the University of Asia Pacific and as a Graduate Head, from 2012 to 2019. His research experience, within both academia and industry, results in over 120 journal and conference publications. His research work contributed to the reputed journal of Scientific Reports (Nature), Knowledge-Based Systems, Artificial Intelligence Review, IEEE ACCESS, Sensors, Cancers, and Applied Sciences. For more than ten years, he has been with the master's and undergraduate students as a supervisor of their thesis work. His research interests include artificial intelligence (AI), machine learning, deep learning, natural language processing (NLP), and big data analysis. He has served as a program committee member in several international conferences/workshops. He served as an Associate Editor for several journals including PLOS ONE journal. He has served as a Reviewer for reputed journals and IEEE conferences like HONET, ICIEV, ICCIT, IJCCI, ICAEE, ICCAIE, ICSIPA, SCORED, ISIEA, APACE, ICOS, ISCAIE, BEIAC, ISWTA, IC3e, ISWTA, CoAST, icIVPR, ICSCT, 3ICT, and DATA21.

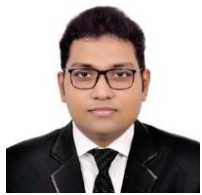

**KHAN MD HASIB** (Member, IEEE) received the B.Sc. degree from the Computer Science and Engineering Department, Ahsanullah University of Science and Technology, Dhaka, Bangladesh, in 2018, and the M.Sc. degree from the Department of Computer Science and Engineering, BRAC University, Dhaka, in 2022. He is currently an Assistant Professor with the Department of Computer Science and Engineering, Bangladesh University of Business and Technology, Dhaka. He has more than four years of teaching and three years of research experience in computer science. He has authored or coauthored over 30 research papers in highly recognized journals, book chapters, and conference proceedings. He is currently working on several projects, such as efficient detection of specific language impairment in children, applied deep learning in image-based cervical cancer detection, activity recognition using Line Follower Robot (LFR), feature-based computerized tomography (CT) image registration of liver cancer, COVID-19 vaccination prediction from numeric data using machine learning algorithms, and Bangla news article classification using explainable artificial intelligence (AI). His research interests include the fields of applied machine learning, medical image processing, health informatics, computer vision, and federated learning.

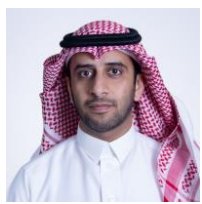

**DR. SULTAN ALFARHOOD** is an Assistant Professor in the Department of Computer Science at King Saud University (KSU). Since joining KSU in 2007, he has made several contributions to the field of computer science through his research and publications. Dr. Alfarhood holds a PhD in Computer Science from the University of Arkansas and has published several research papers on cutting-edge topics such as Machine Learning, Recommender Systems, Linked Open Data, and Text Mining. His work includes proposing innovative approaches and techniques to enhance the accuracy and effectiveness of recommender systems and sentiment analysis.

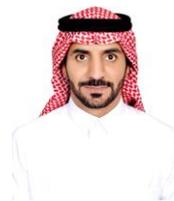

**DR. MEJDL SAFRAN** is an Assistant Professor of Computer Science at King Saud University. He has been a faculty member since 2007 and holds a MSc. (2013) and a PhD (2018) in Computer Science from Southern Illinois University Carbondale. His research interests include computational intelligence, artificial intelligence, deep learning, pattern recognition, and predictive analytics. He has published articles in refereed journals and conference proceedings such as ACM Transactions on Information Systems, Applied Computing and Informatics, MDPI Biomedicine, MDPI Sensors, IEEE International Conference on Cluster, IEEE International Conference on Computer and Information Science, International Conference on Database Systems for Advanced Applications, and International Conference on Computational Science and Computational Intelligence. His current research focus includes developing efficient recommendation algorithms for large-scale systems, predictive models for online human activities, machine learning algorithms for performance management, and modeling and analyzing user behavior. Since 2018, he has been providing part-time consulting services in the field of artificial intelligence to private and public organizations and firms.

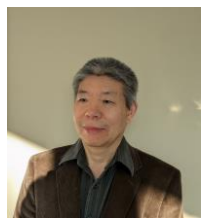

**DR. DUNREN CHE** is currently a professor of Computer Science at Southern Illinois University Carbondale (SIU). He served as the director of undergraduate Computer Science programs in the School of Computing at SIU from 2013-2022. He earned his PhD in Computer Science from Beijing University of Aeronautics and Astronautics in 1994, MS in Computer Science from National University of Defense Technology in 1988, and BS in Electronic Engineering from Harbin University of Commerce (China) in 1985. Before Joined SIU in 2001, he had worked as a post-doctoral research fellow respectively in Tsinghua University, German National Research Center for Information Technology, and Johns Hopkins University. His main research interests are Database, Data Mining, and Machine Learning (collectively Data Science), Cloud Computing, Scientific Workflow. He has authored/co-authored more than 120 peer reviews papers published in various venues such as VLDB Journal, Future Generation Computer Systems, and various ACM/IEEE transactions and associated conferences.

...